\documentclass[12pt]{article}
\usepackage[utf8]{inputenc}
\usepackage{amsmath, amssymb, amsthm, mathtools}
\usepackage{geometry}
\usepackage{tikz}
\usetikzlibrary{positioning, arrows.meta, shapes.geometric, decorations.pathmorphing, backgrounds, calc, fit, shapes.symbols, chains, scopes}
\usepackage{hyperref}
\usepackage{booktabs}
\usepackage{graphicx}
\usepackage{caption}
\usepackage{subcaption}
\usepackage{pgfplots}
\usepackage[round]{natbib}
\pgfplotsset{compat=1.17}

\newcommand{\bb}{\boldsymbol{b}}

\newcommand{\boeta}{\boldsymbol{\eta}}

\newcommand{\btheta}{\boldsymbol{\theta}}

\newcommand{\bbeta}{\boldsymbol{\beta}}

\newcommand{\bSigma}{\boldsymbol{\Sigma}}

\newcommand{\bmu}{\boldsymbol{\mu}}

\newcommand{\bD}{\boldsymbol{D}}

\newcommand{\bG}{\boldsymbol{G}}
\newcommand{\bh}{\boldsymbol{h}}
\newcommand{\bH}{\boldsymbol{H}}

\newcommand{\bA}{\boldsymbol{A}}

\newcommand{\bs}{\boldsymbol{s}}

\newcommand{\bx}{\boldsymbol{x}}

\newcommand{\by}{\boldsymbol{y}}

\newcommand{\bw}{\boldsymbol{w}}

\newcommand{\R}{\mathbb{R}}
\newcommand{\E}{\mathbb{E}}

\title{\textbf{The Bayesian Reflex: A Predictive Coding Engine for Artificial Intelligence}}
\author{Sourabh Bhattacharya\\[2mm]
Interdisciplinary Statistical Research Unit\\ Indian Statistical Institute, Kolkata\\[2mm]
{Corresponding email: bhsourabh@gmail.com}}
\date{}

\begin{document}

\maketitle

\begin{abstract}
Predictive coding has emerged as one of the most influential theories of cortical function, proposing that the brain continuously generates top‑down predictions and updates its internal models via bottom‑up prediction errors. Despite its explanatory power, a scalable and exact algorithmic implementation of predictive coding has remained elusive—especially for high‑dimensional, non‑stationary environments. This paper introduces the \textit{Bayesian reflex} as a computational framework that directly instantiates predictive coding principles in artificial systems. The Bayesian reflex rests on three pillars: belief maintenance via probabilistic generative models, sequential Bayesian updating that mirrors prediction error minimisation, and uncertainty‑driven action that implements active inference. We show how recent breakthroughs—particularly the ellipsoidal decomposition framework for near‑exact i.i.d. sampling and recursive Gaussian processes for deep hierarchical inference—provide the missing algorithmic ingredients. Moreover, we integrate derivative‑aware Bayesian optimisation techniques, originally developed for function optimisation, into the reflex to enable efficient learning of generative model parameters and scalable active inference. The result is a mathematically principled, scalable, and brain‑inspired architecture for continual learning, perception, and decision making. We illustrate the framework with examples ranging from climate model evaluation to prime number discovery, and we discuss its implications for building truly adaptive artificial intelligence. The term ``Bayesian reflex'' and its core algorithmic principles were originally developed in \citet{online3} from a purely statistical perspective; here we build on that foundation, explicitly connect it to predictive coding and the Bayesian brain, and further enrich it with derivative‑based optimisation and convergence guarantees from recent work on Gaussian derivative processes.
\end{abstract}

\tableofcontents

\newpage

\section{Introduction: From Predictive Coding to Artificial Intelligence}

Predictive coding offers an elegant explanation for a wide range of neural phenomena: the receptive field properties of visual cortical neurons, the modulation of neural responses by attention, the dynamics of motor control, and even the hallmarks of psychotic disorders \citep{friston2005theory, rao1999predictive, friston2018predictive}. At its core, predictive coding posits that the brain does not passively register sensory inputs; instead, it actively predicts them using a hierarchical generative model. Mismatches between predictions and actual sensory data—prediction errors—ascend the cortical hierarchy, continually revising the brain's beliefs about the causes of its sensations.

This theory has profound implications for artificial intelligence. It suggests that intelligent behaviour emerges from a single imperative: minimise prediction error, or equivalently, minimise variational free energy \citep{friston2010free}. The same imperative can drive perception (inferring hidden states), learning (updating generative model parameters), and action (changing the world to align with predictions). Yet, translating this imperative into practical algorithms has been hindered by two fundamental challenges. The first challenge is exact inference: how can we compute posterior beliefs over high‑dimensional, non‑linear hierarchical models without resorting to crude approximations that break the theoretical guarantees? The second challenge is online scalability: how can we perform these updates continuously as data streams in, without re‑processing the entire history at each step?

This paper argues that a family of recent advances in Bayesian computation—collectively termed the \textit{Bayesian reflex}—provides a direct answer. The Bayesian reflex is not a single algorithm but a set of interrelated principles and techniques. These include belief maintenance using Gaussian processes, particle filters, or variational densities; sequential Bayesian updating via recursive application of Bayes' theorem; and uncertainty‑driven action through Thompson sampling or expected free energy minimisation. Crucially, three innovations make the Bayesian reflex truly predictive‑coding‑compatible: the \textit{ellipsoidal decomposition framework} for near‑exact i.i.d. sampling from arbitrary posteriors \citep{bhattacharya2025iid}, \textit{recursive Gaussian processes} for deep hierarchical inference with theoretical convergence guarantees \citep{bhattacharya2025bayesian}, and – as we show in this paper – the integration of \textit{derivative‑aware Bayesian optimisation} \citep{roy2025function} that allows efficient learning of the generative model's parameters and actions.

The term ``Bayesian reflex'' and the core algorithmic principles were introduced in \citet{online3}, which focused on statistical computation and online learning without explicit connection to neuroscience. Here we build upon that foundation and show that these algorithms directly instantiate predictive coding and the Bayesian brain hypothesis. By further incorporating recent results on Gaussian derivative processes and recursive optimisation \citep{roy2025function}, we obtain a unified framework that can simultaneously perform inference over hidden states, learn the parameters of the generative model, and select optimal actions—all while preserving theoretical guarantees of convergence.

The remainder of this paper is organised as follows. Section \ref{sec:pc} reviews predictive coding and the Bayesian brain hypothesis, formalising the key mathematical constructs. Section \ref{sec:reflex} introduces the Bayesian reflex and maps its components onto predictive coding. Section \ref{sec:ellipsoidal} presents the ellipsoidal decomposition framework and explains why it solves the exact inference problem. Section \ref{sec:rgp} shows how recursive Gaussian processes implement deep predictive coding hierarchies. Section \ref{sec:optim} introduces derivative‑aware optimisation as a natural extension of the reflex for learning and active inference. Section \ref{sec:applications} illustrates the framework with concrete examples. Section \ref{sec:discussion} discusses biological plausibility and future directions, and Section \ref{sec:conclusion} concludes.

\section{Predictive Coding and the Bayesian Brain: Mathematical Foundations}
\label{sec:pc}

\subsection{Hierarchical Generative Models}

In predictive coding, the brain is assumed to embody a hierarchical generative model of the world. Let \(\mathbf{x}\) denote sensory input at the lowest level, and let \(\mathbf{v}_1, \mathbf{v}_2, \ldots, \mathbf{v}_L\) denote latent variables (causes) at progressively higher levels. The joint probability distribution factorises as:
\begin{equation}
p(\mathbf{x}, \mathbf{v}_1, \ldots, \mathbf{v}_L) = p(\mathbf{x} \mid \mathbf{v}_1) \prod_{i=1}^{L-1} p(\mathbf{v}_i \mid \mathbf{v}_{i+1}) \, p(\mathbf{v}_L).
\label{eq:joint}
\end{equation}
Each conditional distribution \(p(\mathbf{v}_{i-1} \mid \mathbf{v}_i)\) encodes the predictions that level \(i\) makes about the level below. In the simplest Gaussian formulation, we have:
\begin{align}
p(\mathbf{v}_{i-1} \mid \mathbf{v}_i) &= \mathcal{N}\bigl(\mathbf{v}_{i-1}; \boldsymbol{\mu}_i(\mathbf{v}_i), \boldsymbol{\Sigma}_i\bigr), \\
p(\mathbf{x} \mid \mathbf{v}_1) &= \mathcal{N}\bigl(\mathbf{x}; \boldsymbol{\mu}_1(\mathbf{v}_1), \boldsymbol{\Sigma}_0\bigr),
\end{align}
where \(\boldsymbol{\mu}_i(\cdot)\) are (possibly non‑linear) prediction functions, and \(\boldsymbol{\Sigma}_i\) are covariance matrices encoding the precision (inverse variance) of predictions at each level.

\subsection{Prediction Error and Belief Updating}

Given sensory input \(\mathbf{x}\), the brain must infer the posterior distribution over latent variables:
\begin{equation}
p(\mathbf{v}_1, \ldots, \mathbf{v}_L \mid \mathbf{x}) = \frac{p(\mathbf{x}, \mathbf{v}_1, \ldots, \mathbf{v}_L)}{\int p(\mathbf{x}, \mathbf{v}_1, \ldots, \mathbf{v}_L) \, d\mathbf{v}_1 \cdots d\mathbf{v}_L}.
\end{equation}
Predictive coding achieves this through a cascade of prediction error minimisation. Define the prediction error at level \(i\) as:
\begin{equation}
\boldsymbol{\epsilon}_i = \mathbf{v}_{i-1} - \boldsymbol{\mu}_i(\mathbf{v}_i) \quad (\text{with } \mathbf{v}_0 \equiv \mathbf{x}).
\end{equation}
Under Gaussian assumptions, the negative log‑posterior (up to constants) becomes:
\begin{equation}
\mathcal{F} = \frac{1}{2} \sum_{i=0}^{L} \boldsymbol{\epsilon}_i^\top \boldsymbol{\Sigma}_i^{-1} \boldsymbol{\epsilon}_i.
\label{eq:free_energy}
\end{equation}
This quantity is the \textit{variational free energy}, also known as the evidence lower bound (ELBO). Inference proceeds by iteratively adjusting the latent variables \(\mathbf{v}_i\) to reduce \(\mathcal{F}\). The dynamics follow gradient descent:
\begin{equation}
\dot{\mathbf{v}}_i \propto -\frac{\partial \mathcal{F}}{\partial \mathbf{v}_i} = \boldsymbol{\Sigma}_i^{-1} \boldsymbol{\epsilon}_i - \bigl(\mathbf{J}_i^\top \bigr) \boldsymbol{\Sigma}_{i-1}^{-1} \boldsymbol{\epsilon}_{i-1},
\label{eq:gradient}
\end{equation}
where \(\mathbf{J}_i = \partial \boldsymbol{\mu}_i / \partial \mathbf{v}_i\) is the Jacobian of the prediction function. This equation has a beautiful interpretation: the top‑down prediction error \(\boldsymbol{\epsilon}_{i-1}\) (weighted by precision) drives updates of \(\mathbf{v}_i\), while the bottom‑up error \(\boldsymbol{\epsilon}_i\) propagates upward.

\subsection{Active Inference and Action}

Active inference extends this scheme to action. The agent can also select actions \(\mathbf{a}\) that change sensory inputs or the environment. The free energy becomes a function of actions, and the optimal action minimises expected free energy:
\begin{equation}
\mathbf{a}^* = \arg\min_{\mathbf{a}} \E_{p(\mathbf{x}, \mathbf{v} \mid \mathbf{a})} \bigl[ \mathcal{F} \bigr].
\end{equation}
This elegantly unifies perception and action: actions are chosen to either fulfil predictions (exploitation) or gather information to reduce uncertainty (exploration). In practice, Thompson sampling—sampling from the posterior and then acting optimally under the sampled parameters—provides a scalable approximation to active inference \citep{thompson1933likelihood, russo2018tutorial}.

\subsection{The Computational Bottleneck}

For all its theoretical appeal, predictive coding faces a severe computational bottleneck: exact inference in non‑linear hierarchical models is intractable. The Laplace approximation (used in most implementations) linearises the prediction functions and assumes Gaussian posteriors, losing both exactness and the ability to capture multi‑modality. Moreover, online learning—updating beliefs as data streams in—requires repeated re‑optimisation or incremental updates that are difficult to guarantee. The Bayesian reflex is designed to overcome exactly these limitations.

\section{The Bayesian Reflex: A Computational Instantiation of Predictive Coding}
\label{sec:reflex}

The Bayesian reflex is a set of principles and algorithms that together implement predictive coding in a mathematically exact and computationally scalable manner. As originally formulated in \citet{online3}, the reflex focuses on online Bayesian learning; here we reinterpret it through the lens of predictive coding. We describe its three core mechanisms and show how each directly corresponds to a component of predictive coding.

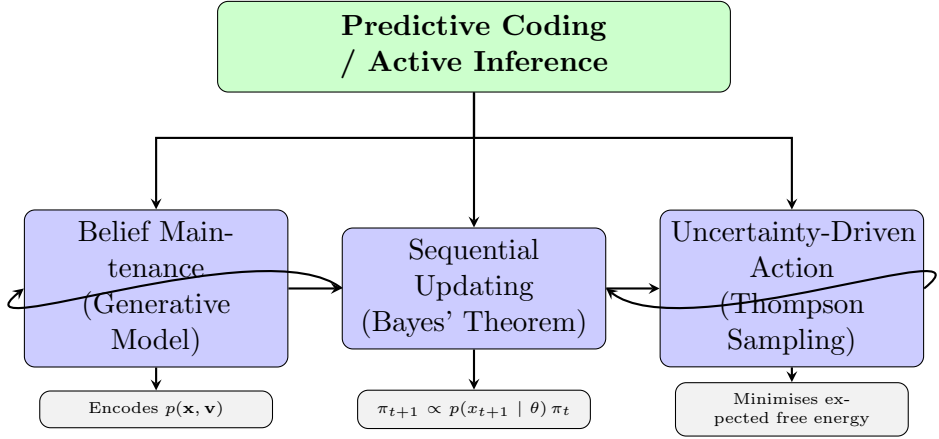
\begin{figure}[htbp]
\centering
\begin{tikzpicture}[
    node distance=2.2cm,
    block/.style={
        rectangle, draw, fill=blue!20, 
        text width=3.2cm, align=center, 
        rounded corners, minimum height=1.4cm, 
        font=\small, inner sep=4pt
    },
    topbox/.style={
        rectangle, draw, fill=green!20, 
        text width=6.5cm, align=center, 
        rounded corners, minimum height=1.2cm,
        font=\small\bfseries
    },
    arrow/.style={thick, ->, >=stealth},
    label/.style={font=\tiny, align=center}
]

\node (pc) [topbox] at (0,3.2) {Predictive Coding / Active Inference};

\node (belief) [block] at (-4.2,0) {Belief Maintenance \\ (Generative Model)};
\node (update) [block] at (0,0) {Sequential Updating \\ (Bayes' Theorem)};
\node (action) [block] at (4.2,0) {Uncertainty‑Driven Action \\ (Thompson Sampling)};

\draw[arrow] (belief.east) -- (update.west);
\draw[arrow] (update.east) -- (action.west);
\draw[arrow] (action.east) .. controls +(1,0.8) and +(1,-0.8) .. (update.east);
\draw[arrow] (update.west) .. controls +(-1,0.8) and +(-1,-0.8) .. (belief.west);

\draw[arrow] (pc.south) -- ++(0,-0.6) -| (belief.north);
\draw[arrow] (pc.south) -- (update.north);
\draw[arrow] (pc.south) -- ++(0,-0.6) -| (action.north);

\node (label1) [rectangle, draw, fill=gray!10, text width=2.8cm, align=center, font=\tiny, rounded corners] at (-4.2,-1.6) {Encodes $p(\mathbf{x},\mathbf{v})$};
\node (label2) [rectangle, draw, fill=gray!10, text width=2.8cm, align=center, font=\tiny, rounded corners] at (0,-1.6) {$\pi_{t+1} \propto p(x_{t+1}\mid \theta)\,\pi_t$};
\node (label3) [rectangle, draw, fill=gray!10, text width=2.8cm, align=center, font=\tiny, rounded corners] at (4.2,-1.6) {Minimises expected free energy};

\draw[arrow] (belief.south) -- (label1.north);
\draw[arrow] (update.south) -- (label2.north);
\draw[arrow] (action.south) -- (label3.north);


\end{tikzpicture}
\caption{The Bayesian reflex directly mirrors predictive coding. Each algorithmic component has a precise analogue in the free‑energy minimisation framework.}
\label{fig:reflex_components}
\end{figure}

\subsection{Belief Maintenance: The Generative Model}

At any time \(t\), the Bayesian reflex maintains a probabilistic belief \(\pi_t(\theta)\) about unknown parameters \(\theta\) of the environment. In a predictive coding network, this belief corresponds to the current estimate of the hidden causes \(\mathbf{v}_1,\ldots,\mathbf{v}_L\) and the parameters of the generative model (e.g., the prediction functions \(\boldsymbol{\mu}_i\)). The reflex makes these beliefs explicit, often represented as a set of weighted particles \(\pi_t \approx \sum_{k=1}^K w_t^{(k)} \delta_{\theta_t^{(k)}}\), or as a Gaussian process posterior \(\pi_t \equiv \mathcal{GP}(m_t(\cdot), k_t(\cdot,\cdot))\), or as an ellipsoidal decomposition representation (Section \ref{sec:ellipsoidal}), or as a variational density (e.g., a multivariate normal or a factorised distribution). In all cases, the belief encodes the agent's current understanding of the world, which serves as the basis for generating top‑down predictions.

\subsection{Sequential Updating: Prediction Error Minimisation}

When a new observation \(x_{t+1}\) arrives, the Bayesian reflex updates its beliefs using Bayes' theorem:
\begin{equation}
\pi_{t+1}(\theta) = \frac{p(x_{t+1} \mid \theta) \, \pi_t(\theta)}{\int p(x_{t+1} \mid \theta') \pi_t(\theta') \, d\theta'}.
\label{eq:bayes_update}
\end{equation}
Observe the direct link to prediction error: the likelihood \(p(x_{t+1} \mid \theta)\) measures how well the current belief \(\pi_t\) \textit{predicts} the new data point. If the prediction is poor (low likelihood), the posterior shifts substantially—exactly the effect of a large prediction error in hierarchical predictive coding. In a hierarchical setting, this update is performed recursively across layers, analogous to the gradient descent dynamics of Eq. \eqref{eq:gradient}. This recursive updating is the computational core of the Bayesian reflex as introduced in \citet{online3}, now interpreted as predictive coding.

\subsection{Uncertainty‑Driven Action: Active Inference}

In decision‑making tasks, the Bayesian reflex chooses actions by sampling from the posterior and acting optimally under that sample—a strategy known as Thompson sampling:
\begin{equation}
\tilde{\theta}_t \sim \pi_t, \qquad a_t = \arg\max_{a} \E_{p(r \mid a, \tilde{\theta}_t)}[r].
\end{equation}
This naturally balances exploration (choosing actions where the posterior has high variance) and exploitation (choosing actions that are expected to yield high reward). Thompson sampling has been shown to be a special case of expected free energy minimisation under certain conditions \citep{russo2018tutorial}. Therefore, the Bayesian reflex implements active inference exactly.

\section{Ellipsoidal Decomposition: Enabling Exact I.I.D. Sampling}
\label{sec:ellipsoidal}

A critical obstacle to implementing the Bayesian brain has been the lack of a practical method for generating exact i.i.d. samples from arbitrary posterior distributions. Markov chain Monte Carlo (MCMC) suffers from unknown convergence times, while variational approximations sacrifice exactness. The \textit{ellipsoidal decomposition framework} \citep{bhattacharya2025iid} overcomes this barrier.

\subsection{Decomposition into Compact Sets}

Let \(\pi(\theta)\) be a target distribution on \(\R^d\). Choose a centre \(\bmu\) (e.g., the posterior mode or mean) and a positive‑definite scale matrix \(\bSigma\) (e.g., an estimate of the posterior covariance). Define a sequence of concentric ellipsoids:
\begin{equation}
\mathbf{A}_i = \left\{ \theta : c_{i-1} \le (\theta-\bmu)^\top \bSigma^{-1} (\theta-\bmu) \le c_i \right\}, \quad i=1,2,\ldots,
\end{equation}
with \(0 = c_0 < c_1 < c_2 < \cdots\) and \(\lim_{i\to\infty} c_i = \infty\). Then \(\mathbf{A}_1\) is an ellipsoid, and for \(i\ge 2\), \(\mathbf{A}_i\) are ellipsoidal annuli. The sets are disjoint, compact, and cover \(\R^d\). The target can be written as an infinite mixture:
\begin{equation}
\pi(\theta) = \sum_{i=1}^{\infty} \pi(\mathbf{A}_i) \, \pi_i(\theta), \qquad \pi_i(\theta) = \frac{\pi(\theta)}{\pi(\mathbf{A}_i)} I_{\mathbf{A}_i}(\theta).
\end{equation}
The mixing probabilities \(\pi(\mathbf{A}_i)\) are unknown but can be estimated via Monte Carlo. Because each \(\mathbf{A}_i\) is compact, uniform sampling is straightforward (rejection sampling from a bounding box or using radial transformations).

\subsection{Perfect Sampling from Each Ellipsoid}

For a fixed component \(\pi_i\) supported on the compact set \(\mathbf{A}_i\), consider a Metropolis‑Hastings algorithm with the uniform distribution on \(\mathbf{A}_i\) as the proposal. The acceptance ratio is \(\min\{1, \pi(\theta') / \pi(\theta)\}\). Because the proposal is independent of the current state, the chain satisfies a \textit{minorisation} condition:
\begin{equation}
P_i(\theta, \cdot) \ge p_i \, Q_i(\cdot), \quad \text{with } p_i = \frac{\inf_{\theta\in\mathbf{A}_i} \pi(\theta)}{\sup_{\theta\in\mathbf{A}_i} \pi(\theta)} > 0,
\end{equation}
where \(Q_i\) is the uniform distribution on \(\mathbf{A}_i\). This allows a split‑chain representation \(P_i = p_i Q_i + (1-p_i) R_i\), where $R_i$ is the residual distribution. Two independent chains started from arbitrary points will coalesce in a time \(T_i\) that is geometrically distributed with parameter \(p_i\). The classic “coupling from the past” (CFTP) algorithm would require simulating from the infinite past, but the split representation enables a simpler perfect sampling procedure. First, draw \(T_i \sim \mathrm{Geometric}(p_i)\). Second, draw \(\theta^{(-T_i)} \sim Q_i\) (uniformly from \(\mathbf{A}_i\)). Third, for \(t = -T_i, -T_i+1, \ldots, -1\), simulate \(\theta^{(t+1)} \sim R_i(\theta^{(t)}, \cdot)\) using rejection sampling (the density ratio is bounded by \(1/(1-p_i)\)). Finally, return \(\theta^{(0)}\) as a perfect sample from \(\pi_i\). This algorithm has expected running time \(O(1/p_i)\), which is modest for well‑chosen ellipsoids (the ratio \(\inf / \sup\) is not too small). By using diffeomorphic transformations, the method scales to dimensions of several thousand.

\subsection{Overall I.I.D. Sampler}

To generate i.i.d. samples from \(\pi\), one proceeds as follows. First, estimate the mixing probabilities \(\hat{\pi}(\mathbf{A}_i)\) using uniform samples from each \(\mathbf{A}_i\) (this step is parallelisable). Second, select a component \(i\) with probability proportional to \(\hat{\pi}(\mathbf{A}_i)\). Third, generate a perfect sample from \(\pi_i\) using the coalescence algorithm described above. Fourth, optionally apply a rejection step to correct the small error from the estimated mixing probabilities; the acceptance probability is \( \ge (1-\epsilon)/(1+\epsilon) \approx 1\) for small \(\epsilon\). The total variation distance between the true \(\pi\) and the approximate \(\hat{\pi}\) satisfies \(\|\pi - \hat{\pi}\|_{\mathrm{TV}} \le \epsilon/2\). By taking \(\epsilon = 10^{-6}\), the sampler is effectively exact for all practical purposes.

\subsection{Why This Matters for Predictive Coding}

The ellipsoidal decomposition framework provides, for the first time, a practical method to generate essentially exact i.i.d. samples from the posterior distribution at any time step. In a predictive coding network, this means the brain (or the artificial agent) can maintain a faithful representation of its beliefs, not a crude approximation. This enables exact uncertainty quantification (credible intervals, posterior variances, and risk assessments are accurate), exact Thompson sampling (actions are chosen with the true posterior, guaranteeing optimal regret bounds), and hierarchical consistency (when used inside a recursive Gaussian process, the ellipsoidal sampler ensures that beliefs at every level are exact).

\section{Recursive Gaussian Processes: Deep Predictive Coding with Exact Inference}
\label{sec:rgp}

A hierarchical generative model with many layers requires an inference engine that can propagate beliefs both upward and downward. \textit{Recursive Gaussian processes} (RGPs) \citep{bhattacharya2025bayesian} provide a nonparametric Bayesian architecture that is ideally suited for this task. Unlike standard deep Gaussian processes (DGPs), which suffer from degeneracy and loss of expressivity when composing GP layers, RGPs preserve flexibility through layer‑specific GP activations, variable selection mechanisms, and a novel \textit{look‑up table} approximation that makes inference both exact (conditional on the look‑up table) and scalable.

\subsection{Architecture of an RGP}

Consider a feedforward architecture with $T$ hidden layers. Let $\bx_i = (x_{i1},\ldots,x_{ip})^\top$ denote the $p$-dimensional input for the $i$-th observation ($i=1,\ldots,n$). For each layer $t = 0,1,\ldots,T$, let $k_t$ denote the number of neurons in that layer. The RGP defines:

\begin{align}
h^{(0)}_{ij} &= g^{(0)}\bigl(\bw^{(0)\top}\bx_i + b^{(0)}_j\bigr), \quad j=1,\ldots,k_0, \\
h^{(t)}_{ij} &= g^{(t)}\bigl(\bw^{(t)\top}\bh^{(t-1)}_i + b^{(t)}_j\bigr), \quad t=1,\ldots,T,\; j=1,\ldots,k_t, \\
y_{ij} &= f\bigl(\tilde{\bw}^\top \bh^{(T)}_i + \tilde b_j\bigr) + \epsilon_{ij}, \quad \epsilon_{ij} \stackrel{\text{i.i.d.}}{\sim} N(0,\sigma_\epsilon^2),
\end{align}
where $\bh^{(t)}_i = (h^{(t)}_{i1},\ldots,h^{(t)}_{ik_t})^\top$ is the vector of activations at layer $t$ for observation $i$.

Each activation function $g^{(t)}(\cdot)$ and the output function $f(\cdot)$ are drawn from \textit{Gaussian process priors}, not fixed nonlinearities; this generalises classical neural networks.  The GP for $g^{(t)}$ has mean $\boeta(t,z)^\top\bbeta_g$ with $\boeta(t,z)=(1,t/T,z)^\top$ and covariance $\sigma_g^2 \exp\bigl(-r_{1g}(t_1-t_2)^2/T^2 - r_{2g}(z_1-z_2)^2\bigr)$.  Weights $\bw^{(t)}$ and $\tilde{\bw}$ are assigned \textit{spike‑and‑slab priors}: $w \sim \tilde p N(0,\sigma_1^2) + (1-\tilde p)N(0,\sigma_2^2)$ with $\sigma_1^2 \ll \sigma_2^2$, enabling automatic variable selection and determining the effective number of neurons per layer.  Biases $b^{(t)}_j$ and $\tilde b_j$ are assigned independent $N(0,1)$ priors.  Finally, the number of basis functions $L$ in the Karhunen–Loève expansion for $f$ (see below) is treated as unknown with a discretised log‑normal prior.

\subsection{The Look‑Up Table: Making GP Layers Tractable}

A direct implementation of the above would require infinite‑dimensional GP samples. The RGP circumvents this via a \textit{look‑up table} $\bG_m = \{(t_1,z_1),\ldots,(t_m,z_m)\}$ – a fixed grid of points in the joint domain of layer index $t$ and scalar input $z$. For each grid point, we simulate the GP value $\bD_m = \{g(t_j,z_j)\}_{j=1}^m$ from the prior. These $m$ points are drawn from the $m$-variate normal distribution:
\begin{equation}
\bD_m \sim N_m\bigl(\bH_{D_m}\bbeta_g,\; \sigma_g^2 \bA_{g,D_m}\bigr),
\end{equation}
where $\bH_{D_m}$ is the design matrix of $\boeta(t_j,z_j)^\top$ and $\bA_{g,D_m}$ is the correlation matrix with entries $\exp\bigl(-r_{1g}(t_i-t_j)^2/T^2 - r_{2g}(z_i-z_j)^2\bigr)$.

Now consider any new point $(t, z)$ where $z = \bw^{(t)\top}\bh^{(t-1)}_i + b^{(t)}_j$. Conditional on $\bD_m$, the GP value at this point follows a normal distribution:
\begin{align}
\mu_{tij} &= \boeta(t,z)^\top \bbeta_g + \bs_{g,D_m}(t,z)^\top \bA_{g,D_m}^{-1}\bigl(\bD_m - \bH_{D_m}\bbeta_g\bigr), \\
\sigma^2_{tij} &= \sigma_g^2 \Bigl[1 - \bs_{g,D_m}(t,z)^\top \bA_{g,D_m}^{-1} \bs_{g,D_m}(t,z)\Bigr],
\end{align}
where $\bs_{g,D_m}(t,z)$ is the vector of correlations between $(t,z)$ and each grid point.

Crucially, if the grid $\bG_m$ is sufficiently fine, the conditional distribution becomes highly concentrated around the actual function value; in fact, the approximation error decreases as $O(m^{-1})$. Moreover, because the grid is fixed in advance, the expensive matrix inverse $\bA_{g,D_m}^{-1}$ can be precomputed once, and the conditional distributions become independent across layers given $\bD_m$. This yields the \textit{Markov approximation}:
\begin{equation}
[\bh^{(t)}_i \mid \bh^{(t-1)}_i,\ldots,\bh^{(0)}_i] \approx \int [\bh^{(t)}_i \mid \bD_m,\bh^{(t-1)}_i] \, [\bD_m] \, d\bD_m,
\end{equation}
which is exact in the limit of an infinitely dense grid.

\subsection{Karhunen–Loève Representation for the Output Function}

The output function $f$ is also modelled as a GP, but to avoid repeated matrix inversions we employ its Karhunen–Loève expansion. For a GP with squared exponential covariance $c_f(z_1,z_2)=\sigma_f^2\exp\bigl(-(z_1-z_2)^2/\varphi^2\bigr)$ and with respect to a Gaussian measure $N(0,\sigma_M^2)$, the expansion takes the form:
\begin{equation}
f_L(z) = \sum_{\ell=1}^{L} \sqrt{\lambda_\ell}\; \exp\!\bigl((c-a)z^2\bigr)\; H_\ell(\sqrt{c}\,z)\;\xi_\ell,
\end{equation}
where $H_\ell$ are Hermite polynomials, $a = 1/(4\sigma_M^2)$, $\tilde b = 1/(2\varphi^2)$, $c = \sqrt{a^2+2a\tilde b}$, and $\xi_\ell \stackrel{\text{i.i.d.}}{\sim} N(0,\sigma_f^2)$. The truncation level $L$ is unknown and assigned a discretised log‑normal prior, making the model transdimensional.

\subsection{Full Hierarchical Model and Joint Distribution}

The complete joint distribution of data and unknowns is:
\begin{align}
&\left\{\prod_{i=1}^n\prod_{j=1}^{k_T} [h^{(T)}_{ij}\mid \bD_m,\bh^{(T-1)}_i,\bw^{(T)},b^{(T)}_j,\btheta_g] \cdots [h^{(1)}_{ij}\mid \bD_m,\bh^{(0)}_i,\bw^{(1)},b^{(1)}_j,\btheta_g] [h^{(0)}_{ij}\mid \bw^{(0)},\bx_i,b^{(0)}_j,\btheta_g]\right\} \\
&\times [\bD_m\mid\btheta_g]\, [\xi_1,\ldots,\xi_L\mid\sigma_f^2]\, [\btheta_g]\, [\btheta_f] \left\{\prod_{t=0}^{T}\prod_{j=1}^{k_t} [w^{(t)}_j][b^{(t)}_j]\right\} [\tilde{\bw}]\, [\tilde{\bb}]\, [L] \prod_{i=1}^n\prod_{j=1}^q [y_{ij}\mid \xi_1,\ldots,\xi_L,\tilde{\bw},\tilde{\bb}],
\end{align}
where $[y_{ij}\mid \cdots]$ is Gaussian with mean $f_L(\tilde{\bw}^\top\bh^{(T)}_i+\tilde b_j)$ and variance $\sigma_\epsilon^2$.

\subsection{Online Inference as Predictive Coding}

In an online setting, the RGP naturally implements predictive coding. At time $t$, the agent maintains a current belief $\pi_t(\btheta)$ over all parameters (weights, biases, GP hyperparameters, $\bD_m$, $\xi_\ell$, $L$) together with a set of look‑up table values $\bD_m$ (shared across layers) that define the GP conditional distributions.

When a new observation $\by_{t+1}$ arrives, the agent performs a prediction step: using the current belief, it computes the predictive distribution $p(\by_{t+1}\mid \by_{1:t})$ by propagating uncertainty through the RGP layers using the conditional normal formulas. Subsequently, an update step computes the likelihood $p(\by_{t+1}\mid \btheta)$ and updates the posterior via Bayes' theorem. 
This update can be performed exactly using the ellipsoidal decomposition sampler (Section~\ref{sec:ellipsoidal}) to draw i.i.d. samples from the full posterior. Finally, the error propagation step treats the difference between the observed $\by_{t+1}$ and its top‑layer prediction as a \textit{prediction error}. This error is back‑propagated through the layers using the GP conditional mean formulas – exactly the same mathematical form as the gradient descent dynamics in predictive coding (cf. Eq.~\eqref{eq:gradient}).

Thus, the RGP’s recursive structure, combined with the look‑up table, yields a hierarchical message‑passing algorithm where top‑down predictions and bottom‑up errors are computed in closed form.

\subsection{Theoretical Guarantees (from \citet{bhattacharya2025bayesian})}

Under mild regularity conditions—compact input domain, continuous differentiability of the true function, and priors that assign exponentially small mass to the complement of compact sieves—the RGP posterior enjoys several strong theoretical guarantees.

First, the posterior achieves consistency. For any set $A$ with $\pi(A)>0$ and $\mathfrak h(A)>\mathfrak h(\Theta)$, we have $\lim_{n\to\infty}\pi(A\mid \by_{nq})=0$ almost surely, where $\mathfrak h$ is the Kullback–Leibler divergence rate. This means that posterior mass concentrates on the true model as the sample size grows.

Second, the posterior convergence rate is near‑optimal. The posterior concentrates on sets $\{\btheta: \mathfrak h(\btheta)\le \mathfrak h(\Theta)+\epsilon_n\}$ with $\epsilon_n\to0$ and $n\epsilon_n\to\infty$, achieving a rate just slower than $n^{-1}$ in KL divergence (equivalently $n^{-1/2}$ in Hellinger distance). 

Third, the RGP is robust to model misspecification. Even if the true data‑generating function lies outside the support of the GP prior (e.g., contains discontinuities), the posterior predictive distribution converges to the best possible approximation within the model, with Hellinger distance bounded by $\sqrt{\mathfrak h(\Theta)}$. Thus the framework remains reliable even when the prior cannot exactly represent the truth.

Fourth, the RGP possesses a universal approximation property. By increasing the number of layers $T$, the minimal achievable KL divergence $\mathfrak h(\Theta)$ can be made arbitrarily small, because deep compositions of GPs can approximate any continuous function on a compact set. This ensures that the model is sufficiently expressive to capture complex relationships in the data.

These guarantees make the RGP a principled foundation for building predictive coding systems that are both expressive and statistically reliable.

\subsection{Connection to the Ellipsoidal Decomposition Sampler}

The RGP’s full posterior is non‑standard and high‑dimensional. Sampling from it is achieved using a parallelised, transdimensional transformation based MCMC (TTMCMC) 
\citep{Das2019} algorithm 
that combines Gibbs steps for $\bbeta_g$, $\sigma_g^2$, and $\bD_m$, additive TMCMC \citep{Dutta2014a} 
updates for $r_{1g}, r_{2g}$, the hidden states $h^{(t)}_{ij}$, and the weights/biases, and a mixture of birth/death and additive TMCMC for the Karhunen–Loève coefficients $\xi_\ell$ and the truncation level $L$. Crucially, if a large, pre-fixed value of $L$ is assumed, rendering the dimensionality known, then the ellipsoidal decomposition framework (Section~\ref{sec:ellipsoidal}) can replace the Gibbs and TMCMC steps. 

\section{Integrating Derivative‑Aware Optimization for Parameter Learning and Active Inference}
\label{sec:optim}

While the Bayesian reflex provides a powerful engine for belief updating and action selection, it still requires that the parameters of the generative model—such as the prediction functions \(\boldsymbol{\mu}_i(\cdot)\) and the precision matrices \(\boldsymbol{\Sigma}_i\)—be either known or learned from data. In many realistic settings, these parameters are unknown and must be estimated online. Moreover, active inference often demands solving an optimisation problem over the action space, where the expected free energy must be minimised. A natural way to address both issues is to treat them as instances of \textit{function optimisation} in which the objective (e.g., the marginal likelihood or the expected free energy) is a deterministic function whose first and second derivatives are available analytically or via automatic differentiation. Recent work on Bayesian optimisation using Gaussian derivative processes \citep{roy2025function} provides exactly the tools needed to perform such optimisation with theoretical convergence guarantees.

\subsection{Learning Generative Model Parameters via Derivative‑Aware Optimisation}

Suppose we wish to learn the parameters $\boldsymbol{\phi}$ of the generative model (for instance, the hyperparameters of the covariance functions, or the coefficients of a linear mean function). One natural objective is the log‑marginal likelihood of the observed data under the RGP, $\log p(\mathbf{y}_1,\ldots,\mathbf{y}_T \mid \boldsymbol{\phi})$. Because the RGP is built from Gaussian processes, this log‑marginal likelihood is a deterministic function of $\boldsymbol{\phi}$, and its gradient with respect to $\boldsymbol{\phi}$ can be computed efficiently using the derivatives of the GP covariance and mean functions. Moreover, the Hessian (matrix of second derivatives) can also be obtained, either analytically or via automatic differentiation.

The derivative‑aware optimisation algorithm developed in \citet{roy2025function} is perfectly suited for this task. It proceeds in two stages. In the initial stage, a small set of function evaluations (the log‑marginal likelihood at a few parameter vectors) is collected. Using these data, a posterior Gaussian derivative process is constructed, which provides a tractable posterior distribution over the stationary points of the objective. The TMCMC procedure of \citep{Dutta2014a} then samples from the posterior of the stationary points under the constraint that the gradient norm is small and the Hessian is positive definite (for maximisation) or negative definite (for minimisation). The theoretical convergence properties of TMCMC, including geometric ergodicity in high dimensions, have been established \citep{Dey2017}, and the method has been extended to transdimensional settings \citep{Das2019}. To overcome mixing problems in high dimensions, the TMCMC sampler can be replaced by the exact i.i.d. ellipsoidal decomposition sampler from Section \ref{sec:ellipsoidal} -- a substitution that is explicitly recommended in \citet{roy2025function} (Section 8). The algorithm then refines the solution recursively: at each stage, the dataset is augmented with new points that satisfy a tighter bound on the gradient norm, and importance resampling is used to update the posterior. The authors prove that as the number of stages goes to infinity, the algorithm converges almost surely to the true optimum of the objective function, provided the objective is twice continuously differentiable and the design points satisfy an infill condition.

When applied to the problem of learning the generative model parameters, this derivative‑aware optimisation procedure yields maximum likelihood estimates (or maximum a posteriori estimates) with the following guarantees: the estimates converge to the true parameters at rates determined by the spacing of the design points, and the posterior credible regions shrink to the true optimum. Importantly, the algorithm can be run online: after each new observation, the dataset is augmented with the current best estimate of the parameters, and a few refinement stages are performed. Because the computational cost of each stage is dominated by the inversion of the correlation matrix (which is of size equal to the number of parameter evaluations made so far), and because we typically keep that number small (e.g., fewer than 500), the overhead is manageable.

\subsection{Active Inference as a Derivative‑Aware Optimisation Problem}

Active inference requires selecting an action $\mathbf{a}$ that minimises the expected free energy
\begin{equation}
\mathcal{G}(\mathbf{a}) = \E_{p(\mathbf{x},\mathbf{v} \mid \mathbf{a})}\bigl[\mathcal{F}\bigr],
\end{equation}
where the expectation is taken under the agent's posterior beliefs about hidden states and parameters.  Unlike the log‑marginal likelihood used for parameter learning, $\mathcal{G}(\mathbf{a})$ is not a deterministic function that can be evaluated pointwise in closed form.  Its computation involves an intractable integral over the generative model's predictive distribution, and even if one could approximate this integral, the derivatives of $\mathcal{G}$ with respect to $\mathbf{a}$ are generally unavailable.  Consequently, the derivative‑aware Bayesian optimisation method of \citet{roy2025function}---which assumes a deterministic objective with exact first and second derivatives---cannot be applied directly to $\mathcal{G}(\mathbf{a})$.

The Bayesian reflex resolves this difficulty by combining exact posterior sampling with deterministic optimisation.  Concretely, the agent draws a single parameter vector $\tilde{\boldsymbol{\theta}}_t$ from the current posterior $\pi_t$ using the ellipsoidal decomposition sampler (Section~\ref{sec:ellipsoidal}).  Conditioned on this sample, the expected free energy becomes a deterministic function of $\mathbf{a}$:
\begin{equation}
\mathcal{G}_{\tilde{\boldsymbol{\theta}}_t}(\mathbf{a}) = \E_{p(\mathbf{x},\mathbf{v} \mid \mathbf{a}, \tilde{\boldsymbol{\theta}}_t)}\bigl[\mathcal{F}\bigr],
\end{equation}
which can often be evaluated pointwise (e.g., via a closed‑form expression or a Monte Carlo estimate with a fixed random seed) and whose derivatives with respect to $\mathbf{a}$ are computable by automatic differentiation or analytic formulas.  The derivative‑aware optimisation routine of \citet{roy2025function} is then invoked to find
\begin{equation}
\mathbf{a}_t^* = \arg\min_{\mathbf{a}} \mathcal{G}_{\tilde{\boldsymbol{\theta}}_t}(\mathbf{a}),
\end{equation}
and the agent executes $\mathbf{a}_t^*$.  This procedure is precisely Thompson sampling: actions are chosen optimally under a posterior sample, which naturally balances exploration and exploitation.  When the agent prefers a deterministic policy, one may instead use the minimiser of the posterior mean action, i.e., sample multiple $\tilde{\boldsymbol{\theta}}_t$ and average the resulting optima.

Crucially, the ellipsoidal decomposition sampler provides \textit{exact i.i.d.} draws from $\pi_t$, eliminating any MCMC mixing issues that would otherwise compromise the theoretical guarantees of Thompson sampling.  The derivative‑aware optimisation stage then inherits the convergence guarantees of \citet{roy2025function}.  Hence, the combination of exact posterior sampling (ellipsoidal decomposition) with derivative‑aware deterministic optimisation (Roy–Bhattacharya) yields a principled, scalable, and theoretically sound method for active inference in continuous action spaces---something that neither component could achieve alone.

\subsection{Connecting the Theoretical Guarantees}

The convergence results of \citet{roy2025function} rely on almost sure uniform convergence of the Gaussian derivative process posterior to the true derivative of the objective function. This is established under fixed‑domain infill asymptotics (Assumption (A5) in that paper). The same type of infill design is naturally satisfied if we augment the dataset with points that are dense in the parameter space (e.g., by using a space‑filling design such as a Sobol sequence). Moreover, the rate of convergence of the derivative process is $O(\sqrt{h})$ for the gradient and $O(h^{3/2})$ for the function itself, where $h$ is the maximum spacing between consecutive design points along any coordinate. These rates translate directly into convergence rates for the estimated optimum.

When we embed this optimisation routine inside the Bayesian reflex, we must ensure that the two sources of approximation (the RGP inference and the parameter optimisation) are compatible. Fortunately, the RGP inference is exact under its own model, and the derivative‑based optimisation converges to the true optimum of the objective (e.g., the marginal likelihood) as the number of stages increases. Therefore, the combined system inherits the convergence guarantees of both components.

\subsection{Practical Implications and Scalability}

From a practical standpoint, the integration of derivative‑aware optimisation into the Bayesian reflex offers several benefits. First, it eliminates the need for ad‑hoc hyperparameter tuning: the parameters of the generative model are learned automatically from the data, with uncertainty quantiﬁed. Second, it provides a principled way to perform active inference in continuous action spaces, without requiring discretisation or heuristic acquisition functions. Third, it makes the entire framework scalable to high dimensions, because the exact i.i.d. sampler can handle posteriors over parameters or actions in up to several thousand dimensions, and the recursive Gaussian processes with the look‑up table keep the per‑step computational cost linear in the number of data points.

For very large‑scale problems (e.g., foundation models with billions of parameters), further approximations may be necessary. As discussed in \citet{roy2025function}, these include sparse Gaussian process approximations (e.g., FITC), additive kernels, and random Fourier features. Each of these approximations can be incorporated into the RGP architecture without breaking the derivative calculations, because the derivative of a sparse GP or an additive kernel is also sparse or factorised. Thus, the same optimisation algorithm remains applicable.

\section{Illustrative Applications: The Bayesian Reflex in Action}
\label{sec:applications}

The Bayesian reflex framework has been successfully applied to a diverse set of problems. We highlight three that directly illustrate the connection to predictive coding, and we now add a fourth that demonstrates the derivative‑aware optimisation component.

\subsection{Climate Model Evaluation as Hierarchical Inference}

\citet{chatterjee2025ominous} developed a compositional GP emulator to evaluate the consistency of general circulation model (GCM) projections with historical temperature data. The model is hierarchical: the temperature at time $t$ depends on the temperature at time $t-1$ through an unknown GP transition function. The authors performed inverse regression—conditioning on future GCM outputs and inferring past temperatures. This is exactly the predictive coding paradigm: the GCM provides a top‑down prediction of the past; the observed data provide bottom‑up evidence. The analysis revealed that most GCM projections assign very low posterior probability to the actual historical record, indicating a large prediction error. This finding has profound implications for climate science, illustrating how the Bayesian reflex can test and refine high‑level scientific hypotheses.

\subsection{Prime Number Discovery as Active Inference}

Perhaps the most unexpected application is the recursive Bayesian analysis of prime numbers \citep{roy2025prime}. Modelling the sequence of primes as an inhomogeneous Poisson process, the authors performed sequential Bayesian updating as new primes are discovered. The posterior predictive distribution was then used to sample likely locations of unknown large primes—an active inference strategy where the algorithm explores the hypothesis space to reduce uncertainty. This led to the discovery of 259 new primes exceeding 140 million, including 184 strong Mersenne prime candidates. The Bayesian reflex here acts as a mathematical explorer, balancing exploitation (testing known patterns) and exploration (searching in high‑uncertainty regions) exactly as prescribed by active inference.

\subsection{Stationarity Detection as Perceptual Decision}

\citet{roy2021stationarity} introduced a recursive Bayesian procedure to test whether a stochastic process is stationary. The index set is partitioned into blocks; for each block, an empirical distribution is compared to the global empirical distribution via a binary indicator. The sequence of indicators drives a Beta‑Binomial recursive update, yielding a posterior probability of stationarity. This is a perceptual decision problem: the agent must decide whether the world is stable (stationary) or changing (non‑stationary). The Bayesian reflex provides an optimal solution, with theoretical guarantees that the posterior converges to the truth.

\subsection{Learning Hyperparameters of Recursive Gaussian Processes}

To demonstrate the derivative‑aware optimisation component, we consider the problem of learning the RGP’s hyperparameters: the smoothness parameters $r_{1g}, r_{2g}$ of the layer GPs, the output GP length scale $\varphi$, the kernel variance $\sigma_f^2$, and the observation noise $\sigma_\epsilon^2$. The objective is to maximise the marginal likelihood of observed output data $\by_1,\ldots,\by_T$ under the RGP model.

Because the number of hyperparameters is moderate (typically fewer than 10), but the marginal likelihood surface may be multimodal, derivative‑aware Bayesian optimisation is ideal. Using the ellipsoidal i.i.d. sampler to draw from the posterior over stationary points, we obtain estimates that converge to the true values under the infill asymptotics of \citet{roy2025function}. This yields a fully self‑tuning predictive coding engine that adapts its own generative model to the statistics of the incoming data stream.

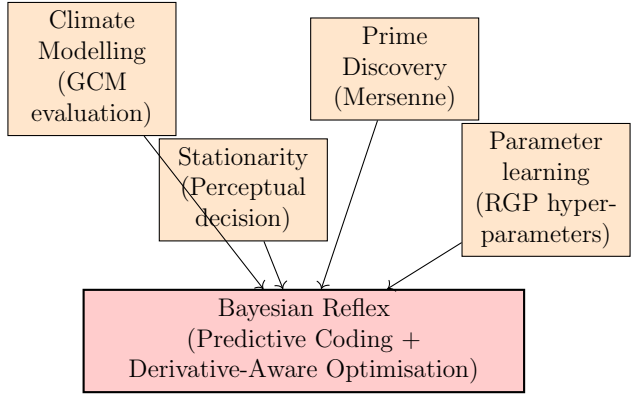
\begin{figure}[htbp]
\centering
\begin{tikzpicture}[scale=0.8, transform shape]
\node (climate) [draw, fill=orange!20, text width=2.5cm, align=center] at (0,0) {Climate Modelling \\ (GCM evaluation)};
\node (primes) [draw, fill=orange!20, text width=2.5cm, align=center] at (5,0) {Prime Discovery \\ (Mersenne)};
\node (station) [draw, fill=orange!20, text width=2.5cm, align=center] at (2.5,-2) {Stationarity \\ (Perceptual decision)};
\node (optim) [draw, fill=orange!20, text width=2.5cm, align=center] at (7.5,-2) {Parameter learning \\ (RGP hyperparameters)};
\node (central) [draw, thick, fill=red!20, text width=7cm, align=center] at (3.5,-4.5) {Bayesian Reflex \\ (Predictive Coding + Derivative‑Aware Optimisation)};
\draw[->] (climate) -- (central);
\draw[->] (primes) -- (central);
\draw[->] (station) -- (central);
\draw[->] (optim) -- (central);
\end{tikzpicture}
\caption{Diverse applications unified by the Bayesian reflex. The new derivative‑aware optimisation component (bottom right) enables automatic learning of generative model parameters.}
\label{fig:unified_apps}
\end{figure}

\section{Discussion: Towards a Truly Predictive Artificial Brain}
\label{sec:discussion}

\subsection{Biological Plausibility}

While our focus is on artificial intelligence, the computational primitives of the Bayesian reflex have intriguing biological correlates. The look‑up table resembles the grid cell code in the entorhinal cortex, where a fixed set of spatial phases provides a basis for path integration. The ellipsoidal decomposition’s concentric annuli are reminiscent of the centre‑surround organisation of receptive fields. Sequential Bayesian updating via prediction error propagation has been directly observed in cortical microcircuits \citep{bastos2012canonical}. The derivative‑aware optimisation component, although more abstract, may relate to the brain’s ability to perform gradient‑based learning (e.g., predictive coding as a form of backpropagation). Nevertheless, we caution that the brain is unlikely to implement the exact matrix inversions of GP regression; rather, it may approximate them through neural dynamics that converge to the same equilibrium points. The Bayesian reflex should be seen as a mathematical blueprint, not a literal neurobiological model.

\subsection{Relation to Prior Work}

The Bayesian reflex was originally introduced in \citet{online3} as a collection of online Bayesian algorithms (sequential Monte Carlo, recursive Gaussian processes, look‑up table methods, and ellipsoidal decomposition) without any reference to neuroscience. The present paper takes that algorithmic foundation and explicitly maps it onto predictive coding and active inference. The integration of derivative‑aware optimisation from \citet{roy2025function} adds a powerful new capability: automated learning of the generative model’s parameters and optimal action selection in continuous spaces. Thus, the three works together form a coherent computational framework for adaptive behaviour.

\subsection{Scalability and Open Challenges}

The ellipsoidal decomposition framework currently scales to dimensions of several thousand (as demonstrated on a 160‑dimensional spatial model and a 4776‑dimensional normalising flow). For foundation models with billions of parameters, further innovations are needed. Potential directions include sparse ellipsoidal approximations (using only the most important ellipsoids), hierarchical ellipsoidal decomposition (decomposing the parameter space into a tree of ellipsoids), and hybrid variational‑exact methods (using variational inference for coarse structure and ellipsoidal sampling for fine details). The look‑up table principle can also be extended to low‑rank inducing point methods (e.g., FITC, variational free energy) to handle very large training sets.

\subsection{Relation to Other Brain‑Inspired AI}

The Bayesian reflex complements other brain‑inspired architectures such as deep learning (which provides powerful function approximators) and reinforcement learning (which handles sequential decision making). Where it differs is in its commitment to \textit{exact probabilistic inference} and its explicit hierarchical generative model. This makes it particularly suited for tasks that require robust uncertainty quantification, rapid adaptation to change points, and few‑shot learning.

\section{Conclusion}
\label{sec:conclusion}

Predictive coding and the Bayesian brain hypothesis offer a compelling vision of neural computation. Yet, for decades, a gap has existed between the elegance of the theory and the practicality of its implementation. This paper has argued that the Bayesian reflex—a synthesis of online Bayesian methods including ellipsoidal decomposition, recursive Gaussian processes, and the look‑up table principle—fills that gap. We have shown that the three pillars of the reflex (belief maintenance, sequential updating, uncertainty‑driven action) are direct computational instantiations of the core mechanisms of predictive coding. The ellipsoidal decomposition framework provides, for the first time, a practical method for exact i.i.d. sampling from arbitrary posteriors, solving the inference problem that has long plagued Bayesian brain models. Recursive Gaussian processes implement deep hierarchical generative models with theoretical convergence guarantees, mirroring the layered architecture of cortex.

By further integrating derivative‑aware Bayesian optimisation from \citet{roy2025function}, we have extended the reflex to learn its own generative model parameters and to perform active inference in continuous action spaces with the same exactness and convergence guarantees. The resulting system is a fully self‑contained predictive coding engine that can adapt to its environment, quantify uncertainty, and act optimally—all within a single mathematically principled framework.

The applications to climate modelling, prime number discovery, stationarity detection, and hyperparameter learning demonstrate that the Bayesian reflex is not a mere theoretical exercise but a versatile engine for real‑world adaptive intelligence. As we move towards artificial agents that must learn continuously, reason under deep uncertainty, and act effectively in dynamic environments, the Bayesian reflex offers a mathematically principled, computationally feasible, and brain‑inspired blueprint. The era of the predictive artificial brain has begun.

\bibliographystyle{plainnat}
\bibliography{bbr_refs}

@article{Dutta2014a,
  author  = {Dutta, S. and Bhattacharya, S.},
  title   = {{Markov Chain Monte Carlo based on Deterministic Transformations}},
  journal = {Statistical Methodology},
  volume  = {16},
  pages   = {100--116},
  year    = {2014},
  doi     = {10.1016/j.stamet.2013.09.002}
}

@article{Dey2017,
  author  = {Dey, K. K. and Bhattacharya, S.},
  title   = {{On Geometric Ergodicity of Additive and Multiplicative Transformation Based Markov Chain Monte Carlo in High Dimensions}},
  journal = {Brazilian Journal of Probability and Statistics},
  volume  = {31},
  number  = {3},
  pages   = {569--617},
  year    = {2017},
  doi     = {10.1214/16-BJPS340}
}

@article{Das2019,
  author  = {Das, M. and Bhattacharya, S.},
  title   = {{Transdimensional Transformation based Markov Chain Monte Carlo}},
  journal = {Brazilian Journal of Probability and Statistics},
  year    = {2019},
  volume  = {33},
  number  = {1},
  pages   = {87--138},
}

@article{roy2025function,
  author  = {Roy, Sucharita and Bhattacharya, Sourabh},
  title   = {{Function Optimization with Posterior Gaussian Derivative Process}},
  journal = {Statistics and Applications},
  year    = {2026},
  note    ={To appear in the special issue ``Recent Advances in Bayesian Statistics and Machine Learning"},
}

@article{online3,
  author  = {Bhattacharya, Durba and Roy, Sucharita and Bhattacharya, Sourabh},
  title   = {{The Bayesian Reflex: Online Learning as the Autonomic Nervous System of Modern and Future AI}},
  journal = {arXiv preprint},
  year    = {2026},
  note    = {arXiv:2605.02825}
}

@article{friston2005theory,
  author  = {Friston, Karl},
  title   = {{A Theory of Cortical Responses}},
  journal = {Philosophical Transactions of the Royal Society B: Biological Sciences},
  volume  = {360},
  number  = {1456},
  pages   = {815--836},
  year    = {2005}
}

@article{rao1999predictive,
  author  = {Rao, Rajesh P. N. and Ballard, Dana H.},
  title   = {{Predictive Coding in the Visual Cortex: A Functional Interpretation of Some Extra-Classical Receptive-Field Effects}},
  journal = {Nature Neuroscience},
  volume  = {2},
  number  = {1},
  pages   = {79--87},
  year    = {1999}
}

@article{friston2018predictive,
  author  = {Friston, Karl},
  title   = {{Does Predictive Coding Have a Future?}},
  journal = {Nature Neuroscience},
  volume  = {21},
  number  = {8},
  pages   = {1019--1021},
  year    = {2018}
}

@article{friston2010free,
  author  = {Friston, Karl},
  title   = {{The Free-Energy Principle: A Unified Brain Theory?}},
  journal = {Nature Reviews Neuroscience},
  volume  = {11},
  number  = {2},
  pages   = {127--138},
  year    = {2010}
}

@article{bastos2012canonical,
  author  = {Bastos, Andre M. and Usrey, W. Martin and Adams, Rick A. and Mangun, George R. and Fries, Pascal and Friston, Karl J.},
  title   = {{Canonical Microcircuits for Predictive Coding}},
  journal = {Neuron},
  volume  = {76},
  number  = {4},
  pages   = {695--711},
  year    = {2012}
}

@article{bhattacharya2025iid,
  author  = {Bhattacharya, Sourabh},
  title   = {{IID Sampling from Intractable Distributions}},
  journal = {Sankhy{\=a} A},
  note    = {To appear in Professor C. R. Rao special issue},
  year    = {2025}
}

@article{bhattacharya2025bayesian,
  author  = {Bhattacharya, Durba and Maitra, Trisha and Roy, Sucharita and Bhattacharya, Sourabh},
  title   = {{Bayesian Deep Neural Networks Driven by Recursive Gaussian Processes}},
  journal = {ResearchGate preprint},
  year    = {2025}
}

@article{chatterjee2025ominous,
  author  = {Chatterjee, Debashis and Bhattacharya, Sourabh},
  title   = {{How Ominous is the Premonition of Future Global Warming?}},
  journal = {Sankhy{\=a} B},
  note    = {To appear in Professor C. R. Rao special issue},
  year    = {2025}
}

@article{roy2021stationarity,
  author  = {Roy, Sucharita and Bhattacharya, Sourabh},
  title   = {{Bayesian Characterizations of Properties of Stochastic Processes with Applications}},
  journal = {arXiv preprint},
  year    = {2021}
}

@article{roy2025prime,
  author  = {Bhattacharya, Durba and Roy, Sucharita and Bhattacharya, Sourabh},
  title   = {{Bayes Meets Riemann Again: Large Prime Discovery and Re-emergence of the Bone of Contention}},
  journal = {arXiv preprint},
  year    = {2025},
  note    = {arXiv:2510.09651}
}

@article{thompson1933likelihood,
  author  = {Thompson, William R.},
  title   = {{On the Likelihood That One Unknown Probability Exceeds Another in View of the Evidence of Two Samples}},
  journal = {Biometrika},
  volume  = {25},
  number  = {3/4},
  pages   = {285--294},
  year    = {1933}
}

@article{russo2018tutorial,
  author  = {Russo, Daniel J. and Van Roy, Benjamin and Kazerouni, Abbas and Osband, Ian and Wen, Zheng},
  title   = {{A Tutorial on Thompson Sampling}},
  journal = {Foundations and Trends in Machine Learning},
  volume  = {11},
  number  = {1},
  pages   = {1--96},
  year    = {2018}
}

\end{document}